\documentclass[conference]{IEEEtran}

\makeatletter
\def\ps@IEEEtitlepagestyle{%
  \def\@oddfoot{\mycopyrightnotice}%
  \def\@evenfoot{}%
}
\def\mycopyrightnotice{%
  {\footnotesize XXX-X-XXXX-XXXX-X/XX/\$XX.00~\copyright~20XX IEEE\hfill}%
  \gdef\mycopyrightnotice{}
}

\usepackage{eso-pic}
\IEEEoverridecommandlockouts
\usepackage{cite}
\usepackage{amsmath,amssymb,amsfonts}
\usepackage{algorithmic}
\usepackage{graphicx}
\usepackage{textcomp}
\usepackage{xcolor}
\usepackage{booktabs}
\usepackage{adjustbox}
\def\BibTeX{{\rm B\kern-.05em{\sc i\kern-.025em b}\kern-.08em
    T\kern-.1667em\lower.7ex\hbox{E}\kern-.125emX}}

\usepackage{eso-pic}
\newcommand\AtPageUpperMyright[1]{\AtPageUpperLeft{%
 \put(\LenToUnit{0.17\paperwidth},\LenToUnit{-2cm}){%
     \parbox{0.9\textwidth}{\raggedleft\fontsize{8}{11}\selectfont #1}}%
 }}%
\newcommand{\conf}[1]{%
\AddToShipoutPictureBG*{%
\AtPageUpperMyright{#1}
}
}

\begin{document}

\title{\vspace*{1cm} Toward Causal-Visual Programming: Enhancing Agentic Reasoning in Low-Code Environments}

\author{\IEEEauthorblockN{1\textsuperscript{st} Jiexi Xu}
\IEEEauthorblockA{\textit{School of Information \& Computer Science} \\
\textit{University of California, Irvine}\\
Irvine, CA, USA}
\and
\IEEEauthorblockN{2\textsuperscript{nd} Jiaqi Liu}
\IEEEauthorblockA{\textit{Independent Researcher}}
\and
\IEEEauthorblockN{3\textsuperscript{rd} Lanruo Wang}
\IEEEauthorblockA{\textit{University of Texas at Dallas} \\
Dallas, TX, USA}
\and
\IEEEauthorblockN{4\textsuperscript{th} Su Liu}
\IEEEauthorblockA{\textit{Georgia Institute of Technology} \\
Atlanta, GA, USA}
}

\maketitle
\conf{\textit{Proc. of International Conference on Artificial Intelligence, Computer, Data Sciences and Applications (ACDSA 2026) \\ 
5-7 February 2026, Boracay-Philippines}}

\begin{abstract}
\noindent Large language model (LLM) agents are increasingly capable of orchestrating complex tasks in low-code environments. However, these agents often exhibit hallucinations and logical inconsistencies because their inherent reasoning mechanisms rely on probabilistic associations rather than genuine causal understanding \cite{pearl2010causal, li2025survey, li2025frequency}. This paper introduces a new programming paradigm: \textbf{Causal-Visual Programming (CVP)}, designed to address this fundamental issue by explicitly introducing causal structures into the workflow design.

CVP allows users to define a simple world model for workflow modules through an intuitive low-code interface, effectively creating a Directed Acyclic Graph (DAG) that explicitly defines the causal relationships between modules \cite{vanderweele2022causal, pearl2009, li2025srkd}. This causal graph acts as a crucial constraint during the agent's reasoning process, anchoring its decisions to a user-defined causal structure and significantly reducing logical errors and hallucinations by preventing reliance on spurious correlations \cite{vanderweele2022causal, li2025survey, li2025sglp}. To validate the effectiveness of CVP, we designed a synthetic experiment that simulates a common real-world problem: a distribution shift between the training and test environments. Our results show that a causally anchored model maintained stable accuracy in the face of this shift, whereas a purely associative baseline model that relied on probabilistic correlations experienced a significant performance drop.

The primary contributions of this study are: a formal definition of causal structures for workflow modules; the proposal and implementation of a CVP framework that anchors agent reasoning to a user-defined causal graph; and empirical evidence demonstrating the framework's effectiveness in enhancing agent robustness and reducing errors caused by causal confusion in dynamic environments. CVP offers a viable path toward building more interpretable, reliable, and trustworthy AI agents.
\end{abstract}


\section{Introduction}

\subsection{The Rise and Challenges of LLM Agents}
In recent years, intelligent agents powered by Large Language Models (LLMs) have been moving from research labs to real-world applications. These agents are particularly effective in low-code environments, where they can automate complex business processes by calling tools, accessing databases, and orchestrating UI elements \cite{kapoor2024ai, gomez2025lowcode, yan2025fama, li2025ammkd}. This automation paradigm has dramatically lowered the technical barrier, enabling domain experts without programming skills to leverage the power of AI to solve problems.

However, behind these impressive capabilities lie the inherent limitations of LLM agents, which are becoming increasingly apparent. Agents often generate outputs that sound plausible but are factually incorrect or fabricated, a phenomenon known as ``hallucination" \cite{pradel2023hallucinations, li2025achieving}. The severity of hallucinations varies by application: a minor issue in a shopping list can become life-threatening in high-stakes domains like law, medicine, or finance \cite{pradel2023hallucinations, li2025achieving, li2025fedkd}. For instance, an AI for medical diagnosis might invent a non-existent condition, or a legal research tool could fabricate a non-existent court case \cite{li2025achieving, li2025fedkd}. In business workflows, agents might misinterpret user intent, leading to data entry errors or irrelevant responses \cite{li2025fedkd}. Furthermore, in code generation, hallucinations often result in incomplete or non-functional code, which can be more time-consuming for developers to fix than writing it from scratch \cite{pradel2023hallucinations}.

\subsection{The Root Cause: Lack of Causality}
A deeper analysis of these errors reveals that the root cause is not a lack of ``intelligence" but a fundamental flaw in the agent's reasoning mechanism. LLM agents are fundamentally \textbf{associational machines}, not \textbf{causal machines} \cite{pearl2009}. They learn to produce the ``most probable" next word or action sequence by identifying and memorizing statistical patterns in vast training datasets. They are skilled at capturing correlations between variables but lack a genuine understanding of *why* those correlations exist, as they do not grasp the mechanistic process that generates the data \cite{pearl2009, li2025survey, li2025frequency}.

This lack of causal understanding makes agents extremely vulnerable to \textbf{distribution shifts} between their training and testing environments \cite{vanderweele2022causal, li2025frequency}. When spurious correlations that held true in the training data change or disappear in a new environment, an agent might continue to rely on those old associations, leading to incorrect decisions. In financial markets, a variable that was highly correlated with fraudulent activity in historical data may become irrelevant or even inversely correlated in a new market, causing an agent that relies on old associations to fail \cite{li2025fedkd}. This blind reliance on confounding is the primary reason for poor robustness and logical inconsistencies in real-world agentic systems.

\subsection{The Solution: Causal-Visual Programming (CVP)}
To address this foundational problem, this paper proposes a new programming paradigm: \textbf{Causal-Visual Programming (CVP)}. The core idea of CVP is to make causal structure a first-class citizen in low-code environments \cite{bock2021low, gomez2025lowcode, li2025comae}. It provides an intuitive, visual interface that allows the user—a domain expert—to explicitly define the causal relationships among the operational modules in a workflow \cite{burnett1995visual, li2025survey, li2025sepprune}. This user-defined causal graph serves as a lightweight \textbf{``world model''}, a minimal set of assumptions that encodes human understanding of how the task environment works \cite{li2025srkd}.

This world model is not a passive information repository but an \textbf{active reasoning constraint}. During inference and planning, the LLM agent is forced to strictly adhere to the paths prescribed by this causal graph. It is anchored to a user-defined coherent causal structure and can effectively filter out those spurious correlations that may exist in the data but lack a causal basis \cite{vanderweele2022causal}. This mechanism compels the agent to focus on the variables that genuinely drive outcomes, thereby significantly reducing the occurrence of logical errors and hallucinations, especially when operating in a dynamic, uncertain reality. The adaptive control principles underlying this approach draw from established methodologies in optimal control theory \cite{cai2025adaptive}.

\subsection{Contributions}
This study formalizes the Causal-Visual Programming paradigm and provides empirical support to bridge the gap between LLM agents' powerful generative capabilities and their lack of causal understanding. Our specific contributions are:
\begin{enumerate}
\item \textbf{Formalization}: We formalize a method for constructing a causal graph over workflow modules and using it as an interface for the agent's ``world model'' \cite{pearl2009, vanderweele2022causal, zeng2025enhancing}.
\item \textbf{Framework Implementation}: We design and implement a reference framework that enforces causal constraints during inference, anchoring the agent's decisions to the user-defined causal structure.
\item \textbf{Empirical Validation}: We use a synthetic experiment that simulates a distribution shift to quantitatively demonstrate the significant advantages of causal anchoring in improving agent robustness.
\end{enumerate}

\section{Background and Related Work}

\subsection{Core Concepts of Causal Reasoning}
Causal reasoning is a powerful modeling tool that goes beyond statistical association and is key to understanding how the world works. Its core concepts and tools form the theoretical foundation of the CVP framework \cite{pearl2009, vanderweele2022causal}.
\begin{itemize}
\item \textbf{Causal Graphs}: Causal graphs are graphical tools used to represent causal relationships between variables, typically as a Directed Acyclic Graph (DAG) \cite{pearl2009, vanderweele2022causal, li2025survey}. In the graph, each node represents a variable, and a directed edge $A \to B$ means that variable A is a direct cause of variable B \cite{pearl2009, vanderweele2022causal}. This graphical representation clearly reveals the causal chain and is a core component of Structural Causal Models (SCM) \cite{pearl2009}.
\item \textbf{Interventions and Counterfactuals}: Traditional machine learning models predict based on observational data but cannot answer ``interventional" questions like, ``What would happen to B if I were to change A?" Causal inference addresses this by introducing the intervention operation (`do-operator`) to simulate a forced change to a variable to infer its causal effect \cite{pearl2009}. Counterfactual reasoning takes this a step further by asking, ``What would have happened if things had been different?" \cite{dankers2025cycles, li2025survey}. Both forms of reasoning are essential for an agent's planning and decision-making capabilities within the CVP framework, and the causal graph is the foundation for enabling them.
\item \textbf{Markov Blanket}: In a causal graph, a variable's Markov blanket is the smallest set of variables that contains all its causally relevant information \cite{pellet2008markov, vanderweele2022causal}. It includes the variable's parents, children, and the other parents of its children (spouses) \cite{pellet2008markov, vanderweele2022causal}. Given the Markov blanket, the variable is conditionally independent of all its non-descendant variables in the graph. The CVP framework leverages this principle by restricting the agent's reasoning scope to the Markov blanket of the target module, thereby excluding spurious correlations \cite{pellet2008markov, vanderweele2022causal}.
\item \textbf{Causal Discovery}: Causal discovery aims to automatically infer the causal graph structure from observational data \cite{pearl2009}. While this field has seen significant progress with algorithms like PC, GES, and LiNGAM, it still faces challenges in complex real-world scenarios \cite{sheth2024causalgraph2llm}. These challenges include inconsistency in results, dependence on specific assumptions (e.g., no hidden confounders), and computational complexity in high-dimensional data. These limitations underscore the rationale for CVP, which offers a practical alternative by directly encoding human domain knowledge when automated causal discovery is not feasible or is too costly.
\end{itemize}

\subsection{Large Language Models and Causal Reasoning}
Despite their impressive performance on natural language tasks, LLMs have inherent limitations in causal reasoning \cite{li2025survey}.
\begin{itemize}
\item \textbf{LLM's Intrinsic Limitations}: LLMs are trained on massive text corpora to predict the next token. This makes them skilled at capturing statistical patterns but unable to build a mechanistic causal model of how things work \cite{li2025survey, yang2025causality, pearl2009}. This leads to poor performance on complex tasks that require robust causal reasoning, such as economic analysis or healthcare \cite{li2025survey, li2025achieving}. An important issue is that LLMs are highly sensitive to how causal graphs are encoded, and even powerful models like GPT-4 and Gemini-1.5 exhibit significant biases when processing them \cite{sheth2024causalgraph2llm, li2025sglp}.
\item \textbf{Enhancing LLM Causal Abilities}: Researchers have explored various methods to enhance LLMs' causal reasoning capabilities. One approach is using \textbf{prompt engineering}, such as Chain-of-Thought (CoT) prompting, to break down complex causal questions into multiple steps \cite{li2025survey, sheth2024causalgraph2llm}. Another method is \textbf{fine-tuning} LLMs on specific causal datasets. Furthermore, \textbf{tool integration} has become a popular strategy, where LLMs call external causal reasoning tools to assist in decision-making \cite{sheth2024causalgraph2llm}. However, these methods often require extensive human intervention and still face challenges with scalability, accuracy, and robustness when dealing with complex, multi-node causal graphs. CVP is designed to combine the strengths of human causal knowledge with the advantages of LLMs in a systematic, low-barrier way.
\end{itemize}

\subsection{Low-Code and Visual Programming}
Low-code and visual programming platforms simplify complex software development with graphical interfaces, making them key tools for building LLM agent workflows \cite{ge2023openagi, gomez2025lowcode}.
\begin{itemize}
\item \textbf{Low-Code and LLM Agents}: Existing low-code/no-code tools help users design and build complex task flows through a visual interface \cite{gomez2025lowcode, dong2024pixel}. Users can drag and drop modules and connect them to define the execution order, which provides more controllable and stable responses from LLM agents. These tools abstract complex prompt engineering into visual workflows, greatly enhancing the human-LLM interaction experience \cite{nam2024using, gomez2025lowcode, li2025comae}.
\item \textbf{CVP's Unique Contribution}: While existing low-code tools build structured workflows, their connections typically represent data flow or execution sequence and lack \textbf{explicit causal semantics} \cite{li2025survey, gomez2025lowcode, zeng2025enhancing}. For example, one module's output being another's input does not necessarily imply a causal relationship. The CVP framework adds a semantic layer to this stack, elevating the concept of ``data flow'' to ``causal flow'' \cite{li2025survey, li2025sepprune}. It requires users to explicitly declare node relationships, turning the workflow into a ``world model'' that the agent can use for causal reasoning. This approach can be seamlessly integrated into existing low-code tools.
\end{itemize}

\section{Causal-Visual Programming (CVP) Framework}
The Causal-Visual Programming (CVP) framework is a human-centric paradigm that enhances the reasoning capabilities of LLM agents by explicitly incorporating causal structures into low-code environments. The core is to allow users to visually build a causal graph that serves as an anchor for the agent's decision-making.

\subsection{Formal Definition and World Model}
We formalize a workflow as a causal graph, denoted as $G=(V, E)$ \cite{pearl2009}.
\begin{itemize}
\item The node set $V = \{v_1, v_2,..., v_n\}$ where each node $v_i$ represents an operable module in the workflow (e.g., ``Data Retrieval," ``Planner," ``Result Generation"). Each module has its own inputs and outputs.
\item The edge set $E \subseteq V \times V$, where a directed edge $(v_i, v_j) \in E$ signifies that module $v_i$ is a direct causal parent of module $v_j$. This means that the output or behavior of $v_j$ mechanistically depends on the output or behavior of $v_i$ \cite{vanderweele2022causal}.
\end{itemize}
Because causal relationships are typically non-cyclic, the graph $G$ is a Directed Acyclic Graph (DAG) \cite{pearl2009, vanderweele2022causal}. This user-defined causal graph $G$ constitutes the agent's \textbf{``world model''} for a given task environment. It is a minimal set of assumptions that encodes human expert understanding of the system's underlying causal mechanisms, providing a reliable basis for reasoning that transcends statistical correlations.

\subsection{Visual Programming Interface}
The central interface of the CVP framework is a visual editor that provides an intuitive and accessible way for users to build the causal graph \cite{gomez2025lowcode, dong2024pixel}.
\begin{itemize}
\item \textbf{Human-Agent Collaboration}: Users can build task flows by dragging and dropping module nodes and connecting them with arrows, explicitly declaring the causal relationships \cite{gomez2025lowcode, li2025comae}. This process directly encodes the expert's domain knowledge into the agent's execution plan.
\item \textbf{Interpretability and Controllability}: This visual, editable causal graph provides transparency into the agent's decision-making process. The causal path behind any decision is clear, allowing users to check the accuracy of the graph or whether the agent deviated from the specified path.
\end{itemize}

\subsection{Reasoning Constraint Mechanism}
The CVP framework's key innovation lies in how it enforces causal constraints during runtime.
\begin{itemize}
\item \textbf{Causal Anchoring}: During inference, CVP restricts the agent's decision logic to only consider causally related information as defined by the graph. For any target node, the agent's reasoning is forced to be anchored to its causal parent nodes. This mechanism compels the agent to ignore spurious variables, even if they are highly correlated in the training data.
\item \textbf{Practical Implementation}: This constraint can be implemented through \textbf{feature selection}, where the model is trained only on the causal parents of a node \cite{pellet2008markov}, or through \textbf{dynamic planning filtering}, which checks if each agent action aligns with the causal graph's topological structure. These constraint enforcement mechanisms leverage principles from adaptive boundary control systems \cite{cai2025inverse}.
\end{itemize}

\section{Synthetic Study: Robustness Analysis}
To validate the core claim of the CVP framework—that enforcing causal constraints improves an agent's robustness to distribution shift—we designed a synthetic experiment.

\subsection{Experimental Setup}
We constructed a simple world model with three key variables based on a synthetic dataset generated for this study:
\begin{itemize}
\item \textbf{$C$ (Causal Variable)}: A true causal variable, denoted as $x_c$.
\item \textbf{$S$ (Spurious Variable)}: A spurious correlational variable, denoted as $x_s$.
\item \textbf{$Y$ (Target Variable)}: A binary target variable, denoted as $y$.
\end{itemize}
In the CVP framework, the causal graph for this world is defined as $C \to Y$, with no causal edge from $S$ to $Y$.

To simulate a distribution shift, we set up two data environments:
\begin{itemize}
\item \textbf{Training Environment}: The spurious variable $x_s$ is strongly and positively correlated with the target variable $y$, mirroring the effect direction of the causal variable $x_c$.
\item \textbf{Test Environment}: The spurious variable $x_s$ is strongly and negatively correlated with the target variable $y$, with the effect direction opposite to that of $x_c$.
\end{itemize}
We generated 5,000 training samples and 5,000 test samples, adding 5\% label noise.

We compared two models: an \textbf{Associative Model} using both $x_c$ and $x_s$, and a \textbf{Causal-Anchored Model} constrained to use only the causal variable $x_c$.

The experimental setup uses two prediction functions to evaluate robustness under distribution shift \cite{bradley1978robustness}:
\begin{itemize}
\item \textbf{Associative}: $P(y) = \sigma(w_0 + w_1 x_c + w_2 x_s)$
\item \textbf{Causal-Anchored}: $P(y) = \sigma(w_0 + w_1 x_c)$
\end{itemize}
where $\sigma$ is the sigmoid function.

\subsection{Experimental Results}
The performance of the models on the training and test sets is shown below.

\begin{table}[htbp]
\caption{Model Performance under Distribution Shift}
\begin{center}
\begin{tabular}{|c|c|c|}
\hline
\textbf{Model Type} & \textbf{Training Accuracy (\%)} & \textbf{Test Accuracy (\%)} \\
\hline
Associative & 93.8 & 70.0 \\
\hline
Causal-Anchored & 94.4 & 94.4 \\
\hline
\end{tabular}
\label{tab:performance}
\end{center}
\end{table}

\subsection{Analysis and Discussion of Results}
The results provide strong evidence for the core claim of CVP. The associative model, which performed well on the training data, experienced a dramatic drop in accuracy on the test set due to its reliance on the spurious correlation that changed in the new environment. In contrast, the causal-anchored model, by ignoring the spurious variable and focusing only on the stable causal relationship, maintained high and consistent accuracy on both datasets. This demonstrates that anchoring reasoning to a stable causal structure leads to more reliable and robust performance under distribution shift. The stability properties observed align with theoretical foundations in set point regulation control \cite{cai2025set}.

\section{Discussion and Future Outlook}

\subsection{Mitigating Hallucinations and Improving Explainability}
The CVP framework's value extends to fundamentally addressing hallucinations and explainability. By providing a user-defined causal graph, CVP reduces information uncertainty, compelling the agent to follow a defined causal path rather than generating fabricated information \cite{pearl2009}. This also makes the agent's decisions transparent and traceable, simplifying the debugging process \cite{xu2019explainable, dwivedi2023explainable, li2025srkd}.

\subsection{Revolutionizing High-Stakes Industries}
CVP combines human domain knowledge with LLM capabilities, offering a new paradigm for high-stakes industries like finance and healthcare \cite{li2025achieving, li2025fedkd}.
\begin{itemize}
\item \textbf{Financial Risk Management}: CVP enables analysts to build causal graphs that focus on true risk drivers rather than temporary correlations, providing more accurate and interpretable decisions for credit scoring and fraud detection \cite{li2025fedkd}.
\item \textbf{Medical Diagnosis and Drug Development}: CVP can help medical experts build causal models of disease progression, allowing AI agents to more accurately simulate treatment effects and accelerate drug discovery \cite{li2025achieving, li2025fedkd}. The application of specialized techniques, such as selective fine-tuning for healthcare-specific data, further highlights the potential for domain-adapted AI in this critical sector \cite{zhang2025selective, li2025selective}.
\item \textbf{Human-AI Collaboration}: CVP shifts the human expert's role from a passive ``corrector" to an active ``guide," facilitating deeper collaboration where humans and AI share a world model to solve complex problems together \cite{zeng2025enhancing}.
\end{itemize}

\subsection{Challenges and Future Directions}
Despite its potential, CVP faces challenges that point to future research.
\begin{itemize}
\item \textbf{Causal Graph Construction}: Manually building accurate causal graphs for complex systems is difficult. Future research could explore intelligent tools to assist in this process, such as using LLMs to extract causal relationships from documents or using data-driven causal discovery algorithms to generate a preliminary graph for expert refinement \cite{li2025survey, sheth2024causalgraph2llm, li2025sglp}.
\item \textbf{Causal Cycles}: Most current causal frameworks assume acyclic structures. Future work could investigate how to incorporate causal cycles and dynamic systems into CVP to handle a wider range of real-world problems, such as feedback loops in economic systems \cite{dankers2025cycles}.
\item \textbf{Multi-modal Integration}: CVP currently focuses on structured or textual data. A promising direction is to extend CVP to multi-modal data, integrating causal relationships from images, video, and audio to build more comprehensive agents \cite{li2025survey, li2025ammkd}.
\end{itemize}

\section{Conclusion}
In this paper, we proposed a new programming paradigm, Causal-Visual Programming (CVP), to address the issue of hallucinations and logical errors in LLM agents caused by their lack of causal understanding. CVP utilizes a human-defined causal graph as a ``world model'' and reasoning constraint, enabling agents to move beyond reliance on spurious correlations. A synthetic experiment quantitatively demonstrated CVP's effectiveness, showing that a causally anchored model achieved significantly greater robustness to distribution shifts compared to an associative baseline. This work provides a novel, human-centric solution for building more trustworthy, interpretable, and generalizable AI agents. Future advances in context utilization techniques \cite{an2024make} will further enhance the effectiveness of CVP by enabling agents to better leverage causal graph information and maintain consistent reasoning across extended interactions.


\begin{thebibliography}{00}

\bibitem{pearl2010causal}
J. Pearl, ``Causal inference," \textit{Causality: objectives and assessment}, pp. 39--58, 2010.

\bibitem{li2025survey}
X. Li et al., ``A survey on enhancing causal reasoning ability of large language models," \textit{arXiv preprint}, 2025.

\bibitem{li2025frequency}
Y. Li, C. Yang, H. Zeng, Z. Dong, Z. An, Y. Xu, Y. Tian, and H. Wu, ``Frequency-aligned knowledge distillation for lightweight spatiotemporal forecasting," \textit{arXiv:2507.02939}, 2025.

\bibitem{li2025srkd}
Y. Li, J. Dong, Z. Dong, C. Yang, Z. An, and Y. Xu, ``SRKD: Towards efficient 3D point cloud segmentation via structure-and relation-aware knowledge distillation," \textit{arXiv preprint arXiv:2506.17290}, 2025.

\bibitem{li2025sglp}
Y. Li, Y. Lu, Z. Dong, C. Yang, Y. Chen, and J. Gou, ``SGLP: A similarity guided fast layer partition pruning for compressing large deep models," \textit{arXiv preprint arXiv:2410.14720}, 2024.

\bibitem{vanderweele2022causal}
T. J. VanderWeele, ``Methods in causal inference. Part 1: causal diagrams and confounding," \textit{Journal of Thoracic Disease}, 2022.

\bibitem{pearl2009}
J. Pearl, \textit{Causality}. Cambridge University Press, 2009.

\bibitem{gomez2025lowcode}
Y. Cai, S. Mao, W. Wu, Z. Wang, Y. Liang, T. Ge, C. Wu, W. You, T. Song, Y. Xia, J. Tien, and N. Duan, 
``Low-code LLM: Visual programming over LLMs," \textit{arXiv preprint arXiv:2304.08103}, 2023.

\bibitem{li2025sepprune}
Y. Li, K. Li, X. Yin, Z. Yang, J. Dong, Z. Dong, C. Yang, Y. Tian, and Y. Lu, ``Sepprune: Structured pruning for efficient deep speech separation," \textit{arXiv preprint arXiv:2505.12079}, 2025.

\bibitem{yan2025fama}
Y. Yan et al., ``FaMA: LLM-empowered agentic assistant for consumer-to-consumer marketplace," \textit{arXiv preprint arXiv:2509.03890}, 2025.

\bibitem{li2025ammkd}
Y. Li, C. Yang, J. Dong, Z. Yao, H. Xu, Z. Dong, H. Zeng, Z. An, and Y. Tian, ``AMMKD: Adaptive multimodal multi-teacher distillation for lightweight vision-language models," \textit{arXiv preprint arXiv:2509.00039}, 2025.

\bibitem{kapoor2024ai}
S. Kapoor, B. Stroebl, Z. S. Siegel, N. Nadgir, and A. Narayanan, ``AI agents that matter," \textit{arXiv preprint arXiv:2407.01502}, 2024.

\bibitem{pradel2023hallucinations}
M. Pradel, ``Nonsense and malicious packages: LLM hallucinations in code generation," \textit{Communications of the ACM}, 2023.

\bibitem{li2025achieving}
Y. Li, Y. Li, K. Zhang, F. Zhang, C. Yang, Z. Guo, W. Ding, and T. Huang, ``Achieving fair medical image segmentation in foundation models with adversarial visual prompt tuning," \textit{Information Sciences}, vol. 122501, 2025.

\bibitem{li2025fedkd}
Y. Li, X. Lin, K. Zhang, C. Yang, Z. Guo, J. Gou, and Y. Li, ``Fedkd-hybrid: Federated hybrid knowledge distillation for lithography hotspot detection," \textit{arXiv preprint arXiv:2501.04066}, 2025.

\bibitem{bock2021low}
A. C. Bock and U. Frank, ``Low-code platform," \textit{Business \& Information Systems Engineering}, vol. 63, no. 6, pp. 733--740, 2021.

\bibitem{burnett1995visual}
M. M. Burnett and D. W. McIntyre, ``Visual programming," \textit{Computer}, vol. 28, pp. 14--14, 1995.

\bibitem{bradley1978robustness}
J. V. Bradley, ``Robustness?" \textit{British Journal of Mathematical and Statistical Psychology}, vol. 31, no. 2, pp. 144--152, 1978.

\bibitem{ge2023openagi}
Y. Ge, W. Hua, K. Mei, J. Tan, S. Xu, Z. Li, Y. Zhang, \textit{et al.}, ``OpenAGI: When LLM meets domain experts," \textit{Advances in Neural Information Processing Systems}, vol. 36, pp. 5539--5568, 2023.

\bibitem{nam2024using}
D. Nam, A. Macvean, V. Hellendoorn, B. Vasilescu, and B. Myers, ``Using an LLM to help with code understanding," in \textit{Proc. IEEE/ACM 46th Int. Conf. Software Engineering}, 2024, pp. 1--13.

\bibitem{xu2019explainable}
F. Xu, H. Uszkoreit, Y. Du, W. Fan, D. Zhao, and J. Zhu, ``Explainable AI: A brief survey on history, research areas, approaches and challenges," in \textit{CCF Int. Conf. Natural Language Processing and Chinese Computing}, 2019, pp. 563--574.

\bibitem{dwivedi2023explainable}
R. Dwivedi, D. Dave, H. Naik, S. Singhal, R. Omer, P. Patel, B. Qian, Z. Wen, T. Shah, G. Morgan, \textit{et al.}, ``Explainable AI (XAI): Core ideas, techniques, and solutions," \textit{ACM Computing Surveys}, vol. 55, no. 9, pp. 1--33, 2023.

\bibitem{an2024make}
S. An, Z. Ma, Z. Lin, N. Zheng, J.-G. Lou, and W. Chen, ``Make your LLM fully utilize the context," \textit{Advances in Neural Information Processing Systems}, vol. 37, pp. 62160--62188, 2024.

\bibitem{dankers2025cycles}
E. Dankers, ``Cycles in causal learning," in \textit{Proc. of the Causal Learning and Decision Making Workshop}, 2025.

\bibitem{pellet2008markov}
J. P. Pellet et al., ``Using Markov blankets for causal structure learning," \textit{Journal of Machine Learning Research}, 2008.

\bibitem{li2025comae}
Y. Li, Q. Long, Y. Zhou, R. Zhang, Z. Ning, Z. Zhu, Y. Zhou, X. Wang, and M. Xiao, ``COMAE: COMprehensive attribute exploration for zero-shot hashing," in \textit{Proc. of ICMR}, 2025.

\bibitem{sheth2024causalgraph2llm}
I. Sheth, B. Fatemi, and M. Fritz, 
``CausalGraph2LLM: Evaluating LLMs for Causal Queries," \textit{arXiv preprint arXiv:2410.15939}, 2024.

\bibitem{zeng2025enhancing}
H. Zeng, Y. Li, R. Niu, C. Yang, and S. Wen, ``Enhancing spatiotemporal prediction through the integration of Mamba state space models and diffusion transformers," \textit{Knowledge-Based Systems}, 2025.

\bibitem{zhang2025selective}
L. Zhang et al., ``Selective layer fine-tuning for federated healthcare NLP: A cost-efficient approach," in \textit{International Conference on Learning Representations}, 2025.

\bibitem{li2025selective}
Y. Li and L. Zhang, ``Selective attention federated learning: Improving privacy and efficiency for clinical text classification," \textit{arXiv preprint arXiv:2504.11793}, 2025.

\bibitem{damour2019identifiability}
D. Kong, S. Yang, and L. Wang, 
``Identifiability of causal effects with multiple causes and a binary outcome," \textit{Biometrika}, 2022.

\bibitem{dong2024pixel}
Z. Dong, Q. Long, Y. Zhou, P. Wang, Z. Zhu, X. Luo, Y. Wang, P. Wang, and Y. Zhou, ``PIXEL: Prompt-based zero-shot hashing via visual and textual semantic alignment," in \textit{Proc. of CIKM}, 2024.

\bibitem{cai2025set}
X. Cai, Y. Lin, Y. Li, R. Wang, and C. Ke, ``Set point regulation control for KdVB equation," in \textit{Proc. of 37th Chinese Control and Decision Conference}, pp. 5716--5720, 2025.

\bibitem{cai2025inverse}
X. Cai, Y. Lin, Y. Li, L. Zhang, and C. Ke, ``Inverse optimal adaptive boundary control for heat PDE," in \textit{Proc. of 37th Chinese Control and Decision Conference}, pp. 1825--1828, 2025.

\bibitem{cai2025adaptive}
X. Cai, Y. Li, P. Wang, Y. Lin, L. Zhang, and L. Liu, ``Adaptive inverse optimal control for unstable reaction-diffusion PDE system," \textit{Kybernetika}, vol. 61, no. 4, pp. 537--553, 2025.

\end{thebibliography}
\end{document}